\documentclass[10pt,twocolumn,letterpaper]{article}

\usepackage{iccv}
\usepackage{times}
\usepackage{epsfig}
\usepackage{graphicx}
\usepackage{amsmath}
\usepackage{amssymb}
\usepackage{balance}
\usepackage{booktabs}
\usepackage{enumitem}

\setenumerate[1]{itemsep=-5pt,partopsep=0pt,topsep=0pt}
\setitemize[1]{itemsep=-2pt,partopsep=-5pt,topsep=-5pt}

\usepackage[pagebackref=true,breaklinks=true,letterpaper=true,colorlinks,bookmarks=false]{hyperref}

\iccvfinalcopy % *** Uncomment this line for the final submission

\ificcvfinal\fi
\begin{document}
 
%%%%%%%%% TITLE
\title{Reprojection R-CNN: A Fast and Accurate Object Detector for 360$^{\circ}$ Images}

\author{Pengyu Zhao\\
Peking University\\
{\tt\small pengyuzhao@pku.edu.cn} 
\and
Ansheng You\\
Peking University\\
{\tt\small youansheng@pku.edu.cn} 
\and
Yuanxing Zhang\\
Peking University\\
{\tt\small longo@pku.edu.cn} 
\and
Jiaying Liu\\
Peking University\\
{\tt\small liujiaying@pku.edu.cn} 
\and
Kaigui Bian\\
Peking University\\
{\tt\small bkg@pku.edu.cn} 
\and
Yunhai Tong\\
Peking University\\
{\tt\small yhtong@pku.edu.cn} 
}

\maketitle

\begin{abstract}
% With the advancement of omnidirectional panoramic technology, 360$^{\circ}$ imagery has become increasingly popular in the past few years.
% 360$^{\circ}$ imagery has been attracting extensive attention for its wide field-of-view.
% Omnidirectional cameras offer great benefits over conventional cameras for the wide field-of-view.
% Detection is fundamental for 360$^{\circ}$ images in real-world applications yet challenging due to the inherent distortion. 
% Several approaches have been recently proposed for the classification and detection tasks in 360$^{\circ}$ images.
360$^{\circ}$ images are usually represented in either equirectangular projection (ERP) or multiple perspective projections.
% ERP displays a omnidirectional view of the 360 image while introducing distortion to the objects in the image.
% Multiple perspective projections can get rid of distortion, whereas a large number of separate perspective projections are needed to compactly cover all objects on the sphere.
Different from the flat 2D images, the detection task is challenging for 360$^{\circ}$ images due to the distortion of ERP and the inefficiency of perspective projections.
However, existing methods mostly focus on one of the above representations instead of both, leading to limited detection performance.
%Moreover, the lack of 360$^{\circ}$ datasets with appropriate bounding-box annotations increases the difficulties of the detection task.
Moreover, the lack of appropriate bounding-box annotations as well as the annotated datasets further increases the difficulties of the detection task.
In this paper, we present a standard object detection framework for 360$^{\circ}$ images.
Specifically, we adapt the terminologies of the traditional object detection task to the omnidirectional scenarios, and propose a novel two-stage object detector, \emph{i.e.}, Reprojection R-CNN by combining both ERP and perspective projection.
%, which takes advantage of both ERP and perspective projections.
%Reprojection R-CNN first generates coarse region proposals by a distortion-aware spherical region proposal network (SphRPN) efficiently owing to the omnidirectional field-of-view of ERP.
Owing to the omnidirectional field-of-view of ERP, Reprojection R-CNN first generates coarse region proposals efficiently by a distortion-aware spherical region proposal network.
Then, it leverages the distortion-free perspective projection and refines the proposed regions by a novel reprojection network.
We construct two novel synthetic datasets for training and evaluation.
Experiments reveal that Reprojection R-CNN outperforms the previous state-of-the-art methods on the mAP metric.
In addition, the proposed detector could run at 178ms per image in the panoramic datasets, which implies its practicability in real-world applications.
\end{abstract}

\section{Introduction}\label{sec:intro}
\begin{figure}[t]
    \centering
    \includegraphics[width=1.0\linewidth]{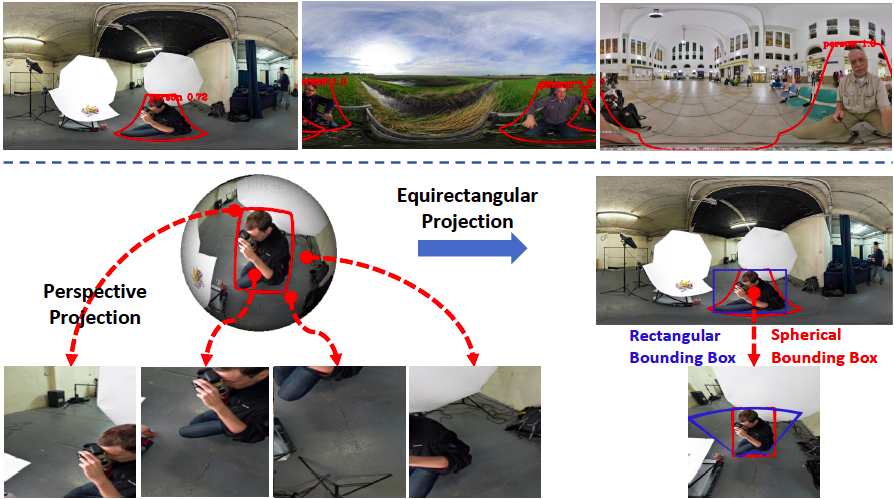}
    \caption{The challenges of the 360$^{\circ}$ object detection task. Objects in ERP suffer from severe distortion as well as discontinuity on the borders. Besides, the object can hardly be recognized with only a few perspective projections. It is also obvious that rectangular bounding box (blue outline) is not appropriate for this task. Despite the challenges, Rep R-CNN is capable of detecting objects in 360$^{\circ}$ images, as shown by the red outlines.}
    \label{fig:teasor}
\end{figure}

% 360 image very hot
% object detection necessary
% but have difficulty
% some work solve, but do not consider the important
% in this work, we ...
% TODO: 要不要碰瓷一下 facebook youtube
During the past few years, virtual reality techniques have developed rapidly owing to the development of $360^{\circ}$ cameras with omnidirectional vision.
The omnidirectional images and videos provide immersive experiences to users, allowing them to receive more detailed information, thereby improving the quality of experiences \cite{ardouin2012flyviz,huang20176}.
$360^{\circ}$ cameras also play important roles in scenarios which require wide-range field-of-view (FoV) such as self-driving systems \cite{chen2018lidar} and gaming \cite{qiu2016unrealcv}.  
%Meanwhile, the growing of convolutional neural networks (CNNs) make the visual tasks easier to be applied on complicated scenarios, and several approaches have been recently proposed to solve classification and detection problems in omni-directional images.
\emph{Object detection} is a significant computer vision task that deals with detecting semantic objects in images and videos. 
Recent advances based on convolutional neural network (CNN)~\cite{girshick2014rich,he2017mask,liu2016ssd,ren2015faster} have achieved remarkable improvements in 2D images.
%Object detection has applications in many areas of computer vision, including image retrieval and video surveillance. 
%For the 360 images, the applications such as pedestrian detection in the self-driving systems, instance recognition in the automated robotics, and interest point detection in the VR videos are also hot topics at the moment.
However, object detection in \emph{360$^{\circ}$ images} (\emph{spherical images}) is still challenging due to two main reasons, as described below and shown in Figure~\ref{fig:teasor}. 

\textbf{Dilemma between Distortion Reduction and Efficiency}. 
Spherical images are typically represented by \emph{equirectangular projection} (ERP) \cite{snyder1997flattening} or multiple \emph{perspective projections}.
Regarding ERP, the coordinates are proportional to latitude and longitude of points on the sphere.
Thus, due to the uneven spatial resolution, ERP suffers from distortions by various latitudes, especially in the polar regions.
Moreover, an object may locate at the borders of ERP and thus be cut into two separate regions, leading to the misinterpretation of two different objects.
Although variants of novel convolutions \cite{coors2018spherenet,tateno2018distortion,zhao2018distortion} are proposed to resolve the distortion problem of ERP, it is still difficult to detect and bound objects in ERP images directly. 
Some researchers resort to perspective projection~\cite{cohen2018spherical,esteves2018learning,lee2018spherephd,yang2018object}, which projects a partial area of the sphere onto a focal plane with little distortion.
However, since we cannot know the exact position of the objects in advance, we need to propose a large number of candidate areas to cover all the objects on the sphere, which is very time-consuming.
%Moreover, distortion still occurs in the at the boundary of the tangent plane, especially when the projection angel is large.
%Although there are some drawbacks in the ERP, the the direct use of the perspective projection is not suitable for the object detection. 
%Owing to the limited field-of-view of the perspective projection, a large amount of perspective images are needed to cover the objects in the sphere, which is similar with the process of the selective search, and is also extremely time-consuming. 
%Furthermore, the objects may be covered by several perspective projections, extracted regions may contain small part of the objects, and the objects may be split by the  is harmful to the object detection. 
%Furthermore, since we do not know the position of the objects, we may sample error regions which contains only a part of the object. 

\textbf{Lack of Appropriate Annotations}. 
% Unlike image classification, extra information is required to localize the objects in the object detection, \emph{i.e.}, bounding box annotations. 
%Unlike image classification task, locating objects in object detection task requires additional information, \emph{i.e.}, bounding boxes.
Unlike image classification task, additional information is required in object detection task to locate objects, \emph{i.e.}, bounding boxes.
% However, the regular bounding box annotations~\cite{yang2018object} do not apply to the $360^{\circ}$ images because a rectangular bounding box in the ERP corresponds to a distorted region in the sphere.
However, the regular bounding boxes~\cite{yang2018object} do not apply to $360^{\circ}$ images because the rectangular area in the ERP corresponds to the twisted region on the sphere.
Recent works~\cite{coors2018spherenet,su2017learning} utilize projection-based annotations that represent the bounding box by the latitude/longitude coordinates of the tangent plane with the object, as well as the width and height of the object on the tangent plane. 
% Although these methods generate distortion-free annotations and work well on certain tasks, they would not accommodate to the scenario where the annotation varies with the center point (viewing angle) and \emph{field of view} (FoV) which cannot be represented on a fixed-size tangent plane. 
% Though the annotation is undistorted, 
%it does not accord with the actual scenes, since the objects on the sphere is not projected from the tangent planes.
% it is task-specified since the annotation varies with the projection ratio between the \emph{field of view} (FoV) of the spherical image and the size of the tangent plane. 
However, 
%In this case, 
it is difficult to measure the intersection-over-union (IoU) between two bounding boxes on a sphere with different center points under this annotation. 
Besides, due to the lack of appropriate annotations, we do not have appropriately annotated datasets for 360$^{\circ}$ object detection.

% TODO: 在某处强调我们先使用 RPN 定位，然后再在这些位置使用 Reprojection mechanism.

To address the above challenges, we introduce an unbiased annotation technique, \emph{i.e.}, \emph{spherical bounding box}, for annotating objects in spherical images.
% In this paper, 
%we introduce a standardized framework for the object detection in the $360^{\circ}$ images.
% we introduce an unbiased annotation technique, \emph{i.e.}, \emph{spherical bounding box}, for annotating objects in spherical images.
Based on the annotation, we propose a novel two-stage object detector, \emph{i.e.}, \emph{Reprojection R-CNN} (Rep R-CNN), for reducing distortion and generating fast object detection in $360^{\circ}$ images.
In the first stage, Rep R-CNN generates coarse detection proposals efficiently from ERP by a novel \emph{region proposal network} (RPN), \emph{i.e.}, \emph{spherical RPN} (SphRPN).
%, which exploits \emph{spherical convolutions} (SphConv) in the network to reduce the distortion.
%to reduce the impact of the distortion in ERP, 
%To reduce the distortion, Rep R-CNN exploits \emph{spherical convolutions} (SphConv) in the network, and extends the conventional \emph{region proposal network} (RPN) to a \emph{spherical RPN} (SphRPN).
%Rep R-CNN extends the conventional \emph{region proposal network} (RPN) to \emph{spherical RPN} (SphRPN) by replacing uniform convolutional filters with \emph{spherical convolutions} (SphConv).
%Rep R-CNN extends the conventional \emph{region proposal network} (RPN) to \emph{spherical RPN} (SphRPN) by replacing uniform convolutional filters with \emph{spherical convolutions} (SphConv).
%, generating coarse detection proposals quickly.
In the second stage, Rep R-CNN utilizes the undistorted perspective projection and introduces \emph{reprojection network} (RepNet) to identify precise locations of the objects.
%Meanwhile, to coordinate the two stages in Rep R-CNN, we also employ a Reprojection Region of Interest (RoI) pooling layer
In addition, a reprojection \emph{region of interest} (RoI) pooling layer is applied to coordinate the two stages in Rep R-CNN.
%we also employ a Reprojection Region of Interest (RoI) pooling layer to eliminate the distortion and coordinate the two stages in Rep R-CNN.
%Due to the lack of annotated 360$^{\circ}$ image datasets, 
We construct two novel datasets, \emph{i.e.}, VOC360 and COCO-Men, for training and evaluation.
Rep R-CNN outperforms all state-of-the-art methods on both datasets, leading to an improvement over the strongest baseline by at least 30\%. 
%We also apply Rep R-CNN on a real-world SUN360 dataset. 
Moreover, the proposed detector could run at about 178ms per frame on an NVIDIA Tesla V100 GPU, indicating that Rep R-CNN is practical in terms of both efficiency and accuracy. 
% practical in terms of both efficiency and accuracy.
%In ablation experiments, we evaluate multiple basic instantiations, which allows us to demonstrate its robustness and analyze the effects of core factors.

%Besides, to standardize the representation for the bounding box, we leverage the spherical FOV to annotate the objects in the $360^{\circ}$ images. Furthermore, the spherical IOU is employed to measure the similarity between the bounding boxes, which accelerates the IOU calculation in the detection process.

%Besides, for the object detection task on the spherical image, we adjust the concepts and standards in the conventional object detection task, and give the annotations for the bounding boxes and propose spherical IOU as the standard metric for the detection task in the spherical images.

In summary, this paper makes the following contributions:
\begin{itemize}
	\item We introduce Rep R-CNN, a fast and accurate two-stage method for object detection in 360$^{\circ}$ images, which takes advantage of both omnidirectional ERP and distortion-free perspective projection.
	\item We adapt the terminologies of conventional object detection task to spherical images, and create two novel synthetic datasets of different scenarios annotated by the proposed spherical bounding box.
	\item We compare Rep R-CNN with several state-of-the-art methods, and demonstrate improved performance on the object detection task of 360$^{\circ}$ images.
\end{itemize}

\section{Related Work}\label{sec:related}
%In this section, we discuss relevant works on $360^{\circ}$ images. 
%\subsection{Methods to reduce distortion in ERP}
%The conventional methods use either use reprojection approach~\cite{su2017making, yu2018deep, yang2018object}, or take direct equirectangular approach~\cite{hu2017deep, deng2017object}. Though the reprojection approach could attain higher accuracy, but it consume a lot of time in reprojection, while the direct CNN approach do not consider the distortion in the spherical images.
%\textbf{CNNs with geometric transformations.}
%To solve the problem of distortion and discontinuities, Khasanova et al.~\cite{khasanova2017graph} propose a graph-based approach, which  represent equirectangular images with weighted graphs, where each image pixel is a vertex and the weights are designed to minimize the difference between filter responses at different image locations. Nevertheless, graph convolutional networks are limited to small graphs, and they do not have apparent performance improvement compared with regular CNNs.
%Another representation, \emph{i.e.}, spherical CNN, that aimed to resolve the problems raised in the ERP representation was proposed in~\cite{cohen2018spherical}. They suggested transforming the domain space from Euclidean S2 space to a SO(3) 3D rotation group to reduce the distortion, and encoding rotation equivariance in the network. 
%However, the full rotation invariance is not desirable since 360 images are mostly captured in one dominant orientation. Besides, it is non-trivial to integrate the spherical CNN to other tasks like object detection.
\noindent \textbf{CNNs on 360$^{\circ}$ Vision}: 
%Recent works design specific CNN models for $360^{\circ}$ images in both representations and architectures. 
%Recent advances in 360$^{\circ}$ images attempt to solve the distortion problem in the ERP.
%Representations other than ERP~\cite{zhang2014panocontext,su2017making,yu2018deep} are introduced in $360^{\circ}$ images.
%Representation based methods other than the ERP or 
%Early methods~\cite{zhang2014panocontext,su2017making,yu2018deep} resort to the accurate perspective projection. However, the projection process is time-consuming.
%~\cite{boomsma2017spherical,cheng2018cube,monroy2018salnet360} replace ERP with Cubemap projection, but introduce additional discontinuities and distortion at the patch boundaries.
%Cubemap projection~\cite{boomsma2017spherical,monroy2018salnet360,cheng2018cube} is utilized in $360^{\circ}$ images.
%Though cubemap projection avoids the distortion varied with latitude in the ERP, it introduces additional discontinuities at the patch boundaries.
%Recently, Lee et al.~\cite{lee2018spherephd} utilizes spherical polyhedron to minimize the variance of spatial resolving power on the sphere surface. However, the pixels in the spherical polyhedrons are over sampled, resulting in a low data efficiency.
% Some others 
Recent advances in 360$^{\circ}$ images 
%attempt to solve the distortion problem
resort to geometric information on the sphere.
%in the $360^{\circ}$ images. 
%Take into account of the geometry in the 360$^{\circ}$ image, 
Khasanova et al.~\cite{khasanova2017graph} represent the ERP with a weighted graph, and apply the graph convolutional network to generate graph-based representations. 
Esteves et al.~\cite{esteves2018learning} propose SO(3) 3D rotation group for retrieval and classification tasks on spherical images.
On top of that, Cohen et al.~\cite{cohen2018spherical} suggest transforming the domain space from Euclidean S2 space to a SO(3) representation to reduce the distortion, and encoding rotation equivariance in the network. 
%However, the above methods are inefficient in practice due to the large time and space overheads. 
%Moreover, the invariance is not desirable in the 360$^{\circ}$ images since they are mostly captured in one dominant orientation
%Meanwhile, novel CNN architectures are designed specifically for $360^{\circ}$ images. 
Meanwhile, some works attempt to solve the distortion in the ERP directly.
%In contrast, architecture based methods introduce novel CNN architectures to solve the distortion in the ERP.
Su et al.~\cite{su2017learning} transfer knowledge from a pre-trained CNN on perspective projections to a novel network on ERP. 
%However, SphConv unties kernel weights along the rows, resulting in a significantly larger number of parameters which makes the model hard to converge.
% The methods of ~introduce a different way of designing CNN for ERP.
Other approaches \cite{coors2018spherenet,tateno2018distortion,zhao2018distortion}  refer to the idea of the deformable convolutional network~\cite{dai2017deformable}, and propose the distortion-aware spherical convolution, where the convolutional filter get distorted in the same way as the objects on the ERP. 
% The convoluted location of SphConv is back-projected from the normal convolutional kernel on the perspective projection to cope with the distortion of objects.
% are same as the grid locations of a normal convolutional kernel back-projected from 
% Thus, SphConv gets distorted in the same way as the objects.
%introducing distortion invariance in the convolutional filters.
Though SphConv is simple and effective, due to the implicit interpolation, it could not eliminate the distortion as the network grows deeper.
To adjust the distortion from SphConv, we introduce a reprojection mechanism in Rep R-CNN, which significantly increases the detection accuracy. 

%They backproject the location of uniform convolutional kernels on the perspective projection at every location on the sphere to the omnidirectional image, and further project the locations to the ERP based on the one-to-one mapping in the polar coordinate. 
%The locations of the convolutional kernels in the ERP thus get distorted in the same way as objects on a perspective projection get distorted when projected from different elevations to an ERP, and introduce distortion invariance in the convolutional filters. 
%Besides, the distortion aware CNN allows the filter sample data across the image boundary which eliminates the discontinuities when processing omnidirectional images with a regular convolutional neural network.
%But the network degrades accuracy when the network grows deeper due to the implicit interpolation assumption. Thus, it could resolve the large image for spherical images perfectly.

%In this paper, we take the distortion aware CNN as the region proposal network.

\noindent\textbf{Object Detection in 2D Images}: 
The promising modern object detectors are usually based on two-stage approaches.
The Region-based CNN (R-CNN) approach~\cite{girshick2014rich} attends to a set of candidate region proposals~\cite{uijlings2013selective} in the first stage, and then uses a convolutional network to regress the bounding boxes and classify the objects in the second stage.
Fast R-CNN~\cite{girshick2015fast} extends R-CNN by extracting the proposals directly on feature maps using RoI pooling.
Faster R-CNN~\cite{ren2015faster} further replaces the slow selective search with a fast region proposal network, achieving improvements on both speed and accuracy.
Numerous extensions have been proposed to this framework~\cite{he2017mask,he2016deep,lin2017feature,shrivastava2016training}.
Compared with two-stage approaches, the single-stage pipeline skips the object proposal stage and generates detection and classification directly, such as SSD~\cite{fu2017dssd,liu2016ssd} and YOLO~\cite{redmon2016you,redmon2017yolo9000,redmon2018yolov3}.
Though these single-stage pipelines attract interests owing to their fast speed, they lack the alignments of the proposals, which is important for 360$^{\circ}$ object detection.
Hence, we adopt the two-stage method in this paper.
%are mostly less accurate than the two-stage detectors. 
%Since the objects in spherical images are difficult to be detected and located, we adopt the accurate two-stage method in this work.

\noindent\textbf{Object Detection in 360$^{\circ}$ Images}:
Object detection in spherical images is an emerging task in computer vision, and several efforts~\cite{coors2018spherenet,su2017learning,yang2018object}  have been made to push forward this issue.
% There are only few worksconsidering the task of object detection in spherical images.
%and they use completely different approaches, datasets and annotations. 
Su et al.~\cite{su2017learning} utilize the network distillation in the network.
% They project the extracted features from ERP to the tangent planes, and apply regular CNN to a specific tangent plane, of which the center is aligned to the object center, to generate region proposals.
This approach applies regular CNN to a specific tangent plane with origin aligned to the object center to generate region proposals.
They construct a synthetic dataset by projecting objects in 2D images onto a sphere. 
Specifically, for each image in the dataset, they select a single bounding box in the image and project it onto the 180th meridian of the sphere with different polar angles.  
%and then compute the IoU on the ERP to evaluate the performance
%Obviously, the approach does not apply in real world object detection task since the position of the bounding box should be unknown. 
Yang et al.~\cite{yang2018object} exploit a perspective-projection based detector on a real-world dataset.
However, they annotate the objects with rectangular regions on ERP, which should have been distorted on the sphere.
%However, the objects are annotated with rectangular regions on ERP, which correspond to distorted regions on the sphere, and thus do not accord with the actual scenarios.
%They select a number of frames in the real-world VR videos and annotate objects  
%The rectangle in the ERP corresponds to a distorted region in the sphere, which is unexpected for the spherical image. 
%Furthermore, the bounding boxes would have a lot of redundancy under the annotation, and could be extremely large in the polar regions. 
%They apply a multi-projection YOLO on the perspective projections, and generate bounding box regions.
%Besides, the applied multi-projection YOLO is also time-consuming for the object detection.
Meanwhile, Coors et al.~\cite{coors2018spherenet} attach the rendered 3D car images to the real-world omnidirectional images and create the synthetic FlyingCars dataset.
% The bounding method is presented in Section~\ref{}.
To solve the distortion in ERP, they utilize the spherical convolution, and apply it to a vanilla SSD.

Apparently, existing methods exploit various settings and examine their performance on synthetic datasets with different annotations.
This situation attributes to the lack of appropriate annotations as well as a standard method for 360$^{\circ}$ object detection.
%Although this situation attributes to the lack of annotations as well as methods, the previous annotations are not appropriate for the 360$^{\circ}$ object detection task.
%Moreover, the dataset annotations are not appropriate for the detection task, and .
Thus, in this paper, we introduce a standardized framework to conform to real-world object detection in 360$^{\circ}$ images, and create two novel datasets for 360$^{\circ}$ object detection task.
% Specifically, we introduce a bounding technique conforming to real-world scenarios, and a two-stage algorithm for object detection task.
% The datasets and source codes for our method would be available soon.

%The bounding boxes of cars are annotated by the latitude and longitude coordinates of the object center, object width and height on the tangent plane and its in-plane rotation. 
%They propose the SphConv and apply SphConv into a vanilla SSD.
%They use ERP IoU to evaluate the algorithm performance.
%They propose a Spherical SSD (Sphere-SSD) which replaces the regular conv with the spherical conv (distortion ware conv), and use spherical anchor boxes to match the bounding boxes in the ERP. 
%Since the bounding boxes could not be evaluated in the tangent plane when the center of them are different, ERP IoU is applied to evaluate the algorithm performance.

\section{Reprojection R-CNN}\label{sec:rep}
In this section, we first establish the new criteria of object detection in 360$^{\circ}$ images.
Beyond that, we present an outline of the proposed two-stage Reprojection R-CNN algorithm, and then introduce the first stage, \emph{i.e.}, Spherical RPN, and the second stage, \emph{i.e.}, reprojection network, respectively.
In the end, we introduce the loss function and the implementation details of the proposed algorithm.

\begin{figure}[t]
    \centering
    \includegraphics[width=1.0\linewidth]{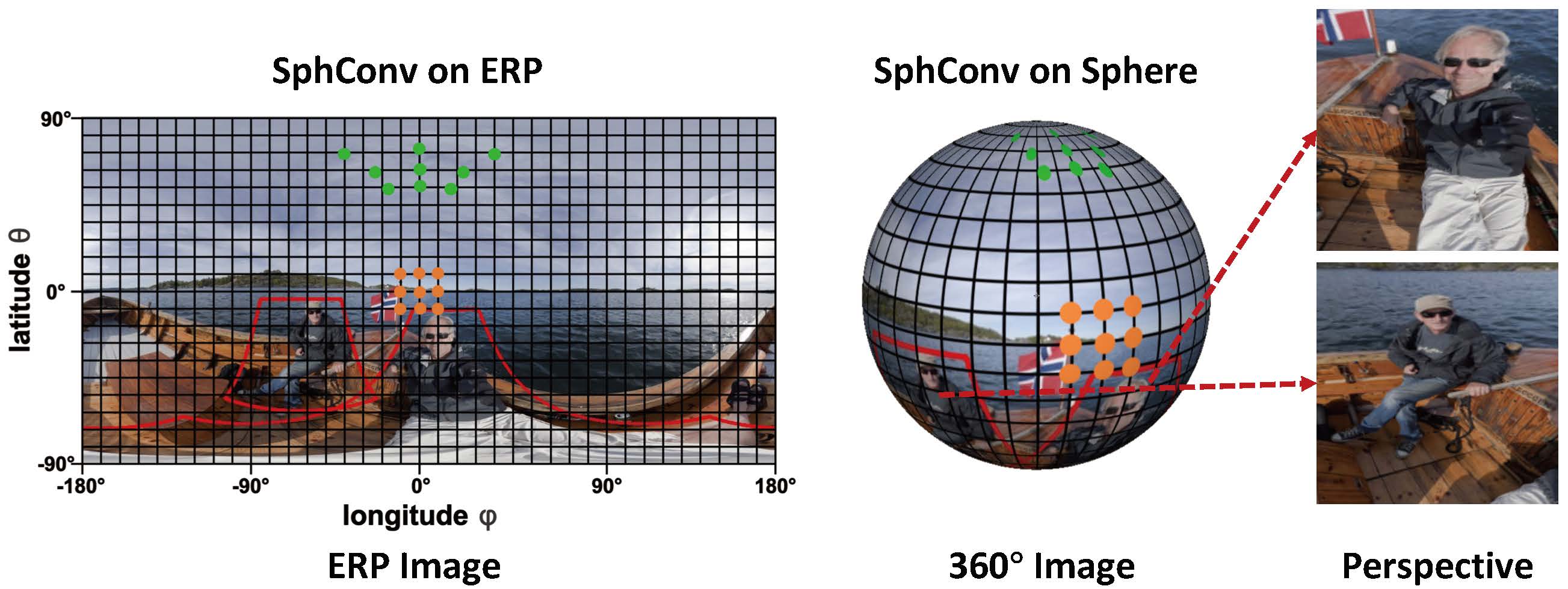}
    \caption{The illustration of the relationship among ERP, sphere, and perspectives, as well as the SphConv used in this paper: The red lines in the ERP delineate the spherical bounding boxes; The green dots and orange dots represent the sampling locations of two spherical convolution kernels at different latitudes.}
    \label{fig:illustration}
\end{figure}

\subsection{Criteria in 360$^{\circ}$ Object Detection}\label{sec:rep_criterion}
%As discussed in Section~\ref{sec:intro} and~\ref{sec:related}, 
The existing bounding-box annotations are not suitable for the object detection task in spherical images.
Thus, we introduce several novel criteria for 360$^{\circ}$ object detection in this subsection, including spherical bounding box, spherical anchor and spherical IoU.
%as well as the novel anchor and IoU 
%\noindent\textbf{Spherical Bounding Box and Spherical IoU}:
% Before going into Rep R-CNN, we will first introduce some criteria for the object detection in spherical image.
%the novel bounding method, \emph{i.e.}, spherical bounding box, for the objects in the 360$^{\circ}$ images.

% TODO: 是不是可以简写一下 fovx fovy，直接变成 x y
\noindent\textbf{Spherical Bounding Box}:
In a real-world omnidirectional scene, if the user's viewpoint coincides with the center of the object, the object will appear with a regular shape centered on the viewpoint, which can be bounded by a spherical rectangle as shown in Figure~\ref{fig:illustration}.
% Following the actual scene, we bound the objects in 360$^{\circ}$ images with the spherical bounding box rather than the distorted rectangular bounding box in ERP. 
Hence, we annotate each object $i$ with a \emph{spherical bounding box}, denoted as $B^i$ with $B^i=(B_{\theta}^i, B_{\phi}^i, B_{fov_x}^i, B_{fov_y}^i)$, where $B_{\theta}^i$ and $B_{\phi}^i$ represent the latitude/longitude coordinates of the object's center (viewpoint), and $B_{fov_x}^i$, $B_{fov_y}^i$ represent the left-right/up-down field-of-view angles of the object's occupation. 
Meanwhile, we also exploit the \emph{spherical bounding-box regression} to tailor the annotations by replacing the height and width in the conventional bounding-box regression with FoVs.

%\noindent\textbf{Spherical Bounding Box Regression}:
% TODO: 可以使用加 Mod 的方法来算 regression,实现的话使用分段函数
%Though the annotation changed, we introduce a similar spherical bounding box regression as~\cite{girshick2014rich}, except that the height and the width of the object are changed by the FoVs. 

\noindent\textbf{Spherical Anchor}:
%Similar with the anchor box in~\cite{ren2015faster}, we use 3 scales and 3 aspect ratios, yielding 9 anchors at each sliding position of ERP. 
Since the proposed spherical bounding box delineates a spherical rectangle, the scales of the anchor boxes in RPN should be measured by FoV angles rather than pixel sizes. 
Therefore, we introduce the translation-invariant \emph{spherical anchors} for 360$^{\circ}$ images, which are represented by the left-right and up-down FoV angles at each sliding-window location (\emph{e.g.}, $30^{\circ} \times 60^{\circ}$).

\noindent\textbf{Spherical IoU}: 
Due to the pixel-wise integral on the sphere, the computation of actual IoU is time-consuming.
Moreover, the existing ERPIoU~\cite{coors2018spherenet,su2017learning} is biased since the pixel size on the sphere varies with longitude, while it still requires the time-consuming pixel-wise calculation on ERP. 
%Moreover, the pixel-wise calculation of ERPIoU and actual IoU on sphere is time-consuming.
%Besides, the computation of the actual IoU is also time-consuming since 
%We find that the calculation of the existing ERPIoU is time-consuming. 
Thus, we introduce a fast and efficient IoU metric based on the proposed spherical bounding box,
%Based on the proposed spherical bounding box, we introduce a fast and efficient IoU metric, 
named as \emph{spherical IoU} (SphIoU).
SphIoU approximately measures the similarity between the spherical bounding boxes, and could be calculated in parallel. 
%Assume that the intersection between two spherical bounding boxes $B^i=(B_{\theta}^i, B_{\phi}^i, B_{fov_x}^i, B_{fov_y}^i)$ and $B^j=(B_{\theta}^j, B_{\phi}^j, B_{fov_x}^j, B_{fov_y}^j)$ form a spherical rectangle on the sphere.
Specifically, SphIoU assumes that the intersection between two spherical bounding boxes $B^i$ and $B^j$ form a spherical rectangle.
The FoV angles of the intersection can then be derived from the difference between the upper left and lower right corners of the spherical rectangle, which is similar to the normal IoU calculation, except that the angle is now represented by polar coordinates, and the width and height are determined by FoV angles.
Meanwhile, the area of a spherical bounding box $B$ with FoV angles of $B_{fov_x}$ and $B_{fov_y}$ on the unit ball can be calculated by:
\begin{equation}
	\text{Area}(B) = 2 B_{fov_x} \, \sin( B_{fov_y} / 2).
\end{equation}
Therefore, the SphIoU between $B^i$ and $B^j$ follows:
\begin{equation}
\begin{small}
	\text{SphIoU}(B^i, B^j) = \frac{\text{Area}(B^i \cap B^j)}{\text{Area}(B^i) + \text{Area}(B^j) - \text{Area}(B^i \cap B^j)}.
\end{small}
\end{equation}

Note that the centers of $B^i$ and $B^j$ may appear in separate boundaries of ERP, and cause wrong gradient descent during training the object detection model. 
Hence, we add both $B_{\phi}^i$ and $B_{\phi}^j$ by 180$^{\circ}$, and obtain a modified SphIoU. 
We take the maximum of the origin SphIoU and the modified SphIoU as the final outcome.
%In addition, 
In another case
when the spherical bounding box covers one pole of the sphere, we cut the spherical bounding box into two sub-regions along the circle of longitude under the assumption that the spherical bounding box is always orthogonal to the latitude line.
We take the sum of the SphIoU between the sub-regions belonging to different spherical bounding boxes as the overall SphIoU.
%Thus, we obtain another SphIoU by adding both latitudes by $\pi$, and take the maximum of both SphIoUs as the final outcome.
%SphIoU enables parallel processing in the calculation, which is much more efficient and faster than the previous ERPIoU.
A further discussion about the computation and approximation of SphIoU is given in the supplementary material.

\begin{figure*}[t]
    \centering
    \includegraphics[width=1.0\linewidth]{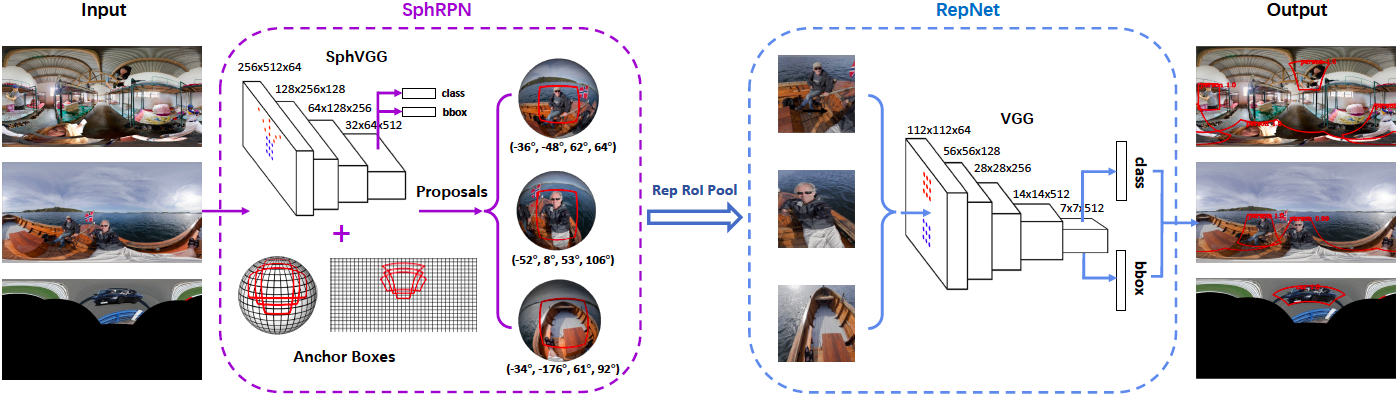}
    \caption{The architecture of the proposed Reprojection R-CNN: The first stage employs SphVGG backbone to generate coarse proposals; The second stage applies a standard VGG backbone to retrieve feature maps for yielding spherical bounding boxes (``bbox'', for short) and objectiveness scores. A reprojection RoI pooling (``Rep RoI Pool'', for short) layer is applied to bridge SphRPN and RepNet. }
    \label{fig:architecture}
\end{figure*}

\subsection{Overview of Reprojection R-CNN}\label{sec:rep_overview}
\noindent\textbf{Combining ERP and Perspective Projection}:
We combine the advantages of two typical representations in spherical images, \emph{i.e.}, ERP and perspective projection, for 360$^{\circ}$ object detection.
Though ERP introduces severe distortion in the image, the ERP-based methods could generate region proposals owing to its omnidirectional view.
%The ERP-based methods could generate region proposals owing to the omnidirectional view of ERP, while ERP introduces severe distortion in the image.
Meanwhile, the perspective projection could eliminate the distortion in the ERP, but a large number of regions are required to cover all the objects in the image due to its limited field-of-view.
%Meanwhile, though the perspective projection could eliminate the distortion in the ERP, a large number of regions are required to cover all the objects in the image due to its limited field-of-view.
% Previous methods~\cite{cohen2018spherical,su2017learning,yang2018object} adopt either representation in the object detection, thus the detector would be either inaccurate or extremely slow.
%In contrast to those methods, 
Unlike previous methods~\cite{cohen2018spherical,su2017learning,yang2018object} adopting single representation in the object detection, we take advantage of both representations in the proposed Rep R-CNN, leading to a fast and accurate object detector for 360$^{\circ}$ images.

%It could extract proposal regions from the omnidirectional view of the ERP, and then relocate the bounding boxes based on the more precise perspective projection. 

\noindent\textbf{Reprojection R-CNN Architecture}:
The overall architecture of Rep R-CNN is illustrated in Figure~\ref{fig:architecture}.
Rep R-CNN contains two stages, where the first stage is a spherical RPN, and the second stage is a reprojection network. 
The SphRPN could propose coarse object detections rapidly on the ERP, while RepNet could refine the proposed regions based on the perspective projections and generate precise spherical bounding boxes.
% for 360$^{\circ}$ images.
%, resulting in precise bounding boxes for the image.
%Referring to the spirit of the Faster R-CNN, we apply the two-stage Reprojection R-CNN as the object detector for 360$^{\circ}$ images. 

%In the first stage, SphRPN takes ERP of size $H_e \times W_e$ as input, and generates coarse region proposal.
Specifically, SphRPN exploits SphConv in the backbone network, and extracts a distortion-aware feature map efficiently from ERP of size $H_e \times W_e$.
%classifies and regresses bounding boxes with reference to anchor boxes of multiple scales and aspect ratios
%In the first stage, SphRPN exploits SphConv in the backbone network, and extracts a distortion-aware feature map from ERP.
A spatial window is slid over the extracted feature map, where at each location SphRPN predicts $k$ spherical bounding boxes based on the corresponding spherical anchors, as well as the objectness score for each proposal.
%Then, a $3 \times 3$ spatial window is slid over the extracted feature map, generating $k$ proposal regions with objectness scores for each location based on the $k$ spherical anchor boxes.
%To generate more precise detections, Rep R-CNN utilizes the perspective projection in the second-stage RepNet.
To bridge SphRPN and RepNet, a reprojection RoI pooling layer is applied to transform the proposed regions to fixed-size perspective projections.
In particular, it expands the region proposals from SphRPN, reprojects the expansions to the tangent planes, and resizes the projected areas to a fixed size of $H_p \times W_p$.
%A Reprojection RoI pooling layer is first employed to bridge the two stages, which expands the proposed regions, reprojects the expanded regions to the tangent planes, and resizes the projected areas to fixed-size feature maps of $H_p \times W_p$.
Then, RepNet takes the distortion-free projections as input and utilizes another backbone network to rectify the spherical bounding box of each projection, generating the final detections of the spherical image.
%The projections are then fed into RepNet with another backbone network, generating the final detection proposals for the spherical image.
%In the second stage, RepNet utilizes the perspective projection and introduces a novel reprojection-based mechanism to relocate the proposals generated from the first stage. 
%We employ a Reprojection RoI pooling layer to coordinate the two stages, .
%The feature maps are then fed into another backbone network in the second stage, generating the final detection proposals for the spherical image.
%Finally, RepNet outputs the classification scores 
%over the $K+1$ categories 
%and the bounding-box regression offsets for each of the object classes.
%for each of the $K$ object classes. 

%Thus, Rep R-CNN generate the proposal regions from ERP, and then relocate the proposals by the more accurate perspective projection.

%In the second stage performs classification and bounding-box regression based on the candidate boxes. 
%Different with the previous detectors, we take advantage of the geometry of the sphere, and solve the distortion in the 360$^{\circ} $ images.
%Specifically, we replace the vanilla RPN with the SphRPN and propose a novel reprojection-based mechanism in the second stage, leading to a fast and accurate object detection in the 360$^{\circ}$ images.

\subsection{Spherical Region Proposal Network}\label{sec:rep_sphrpn}
% SphRPN has the same effect as the vanilla RPN. 
The role of SphRPN is the same as that of vanilla RPN.
Given the ERP of a 360$^{\circ}$ image, SphRPN generates the objectness score, and the offset of the spherical bounding box for each candidate region. 
The only difference is that SphRPN adopts the SphConv~\cite{coors2018spherenet} in the CNN, and introduces the novel spherical anchors as regression reference.
% TODO Figure is needed!

\noindent\textbf{Spherical Convolution}:
The SphConv is designed to address the distortion in ERP. 
It adjusts the sampling locations of the convolutional filter by projecting a uniform convolutional filter on the tangent planes centered at the corresponding locations back to the ERP, as shown in Figure~\ref{fig:illustration}.
%and thus makes the convolutional filters distorted in the same way as the objects on the ERP.
%and encodes invariance into convolutional neural networks.

Formally, assume that the spherical input image is defined on polar coordinates, and the size of ERP is $H_e \times W_e$.
%and $I_e \in \mathbb{R}^{h \times w \times 3}$ be the ERP of the image. 
%Note that there is a one-to-one mapping between the points in the spherical image and the points in the ERP. 
For convenience, we use latitude and longitude, \emph{i.e.}, $\theta \in [-90^{\circ}, 90^{\circ}]$ and $\phi \in [-180^{\circ}, 180^{\circ}]$, to represent the points in ERP.
%hus we may use either the polar coordinates or the rectangular coordinates to represent the point on the ERP. 
Let $I_p[\theta,\phi]$ denote the tangent plane centered at $(\theta, \phi)$ on the sphere,
%generated by the perspective projection, 
where the coordinate system of $I_p[\theta,\phi]$ takes the center as the origin and orients upright.
We only consider the $3 \times 3$ SphConv, since it is sufficient for the backbone network~\cite{simonyan2014very}. 

The sampling locations vary with latitudes to cope with the distortion.
To find the exact sampling locations, we resort to two-step projection transformations.
Specifically, we first sample the locations 
%$s_{i,j}$ with $i,j \in \{-1,0,1\}$ 
of a regular $3 \times 3$ convolutional filter located at the center of ERP, \emph{i.e.}, the coordinate (0,0), with pixel size $\Delta_{\theta} = 180^{\circ} / H_e$ and $\Delta_{\phi} = 360^{\circ} / W_e$.
The sampling locations on $I_p[0,0]$, \emph{i.e.}, ($x_{q,q'}$, $y_{q,q'}$), with $q,q' \in \{-1,0,1\}$, can be calculated via gnomonic projection~\cite{frederick1990map}.
%, \emph{i.e.}, (0,0):
%\begin{align}
%	s_{(0,0)} &= (0,0) \\
%	s_{(\pm1, 0)} &= (\pm \Delta_{\theta}, 0) \\
%	s_{(0, \pm1)} &= (0, \pm \Delta_{\phi})	\\
%	s_{(\pm1, \pm1)} &= (\pm \Delta_{\phi}, \pm \Delta_{\theta})	
%\end{align}
%where $\Delta_{\phi} = \pi / h$ and $\Delta_{\theta} = 2\pi / w$ represent the unit pixel size for the polar coornates, and the subscripts $\pm 1$ and $0$ represent the relative coordinate of the convolutional filter.
%Based on the gnomonic projection~\cite{frederick1990map}, the corresponding locations on $I_p[0,0]$, \emph{i.e.}, $x_{i,j}$, could be represented by:
%The corresponding positions on $I_p[0,0]$, \emph{i.e.}, $x_{i,j}$, could be calculated by gnomonic projection~\cite{frederick1990map}:
%\begin{align}
%	x_{(0,0)} &= (0,0) \\
%	x_{(\pm1, 0)} &= (\pm tan \Delta_{\theta}, 0) \\
%	x_{(0, \pm1)} &= (0, \pm tan \Delta_{\phi})	\\
%	x_{(\pm1, \pm1)} &= (\pm tan \Delta_{\theta}, \pm sec\Delta_{\theta} \, tan \Delta_{\phi})
%\end{align}
% We fix the sampling distribution as the same for each location on the sphere.
%nd copy them to tangent planes of all locations on the sphere. 
Then, we could get the sampling locations of $3 \times 3$ SphConv corresponding to each location on ERP by the inverse gnomonic projection ($f_{\theta}$ and $f_{\phi}$). 
In concrete, the sampling locations on the tangent planes are fixed, and for the SphConv located at $(\theta, \phi)$, we have:
\[
\begin{small}
	f_\theta(x_{(q,q')},y_{(q,q')}) = \text{asin}(\cos \nu \, \sin\theta + \frac{y_{(q,q')} \, \sin\nu \, \cos \theta}{\rho}),
\end{small}
\]
\[
\begin{small}
\begin{aligned}
	f_\phi(x_{(q,q')},y_{(q,q')}) = &\phi + \text{atan}(\frac{x_{(q,q')} \, \sin\nu}{\rho \, \cos \theta \, \cos \nu - y_{(q,q')} \, \sin\theta \, \sin \nu}), 
\end{aligned}
\end{small}
\]
% xxx atan()?
where $\rho = \sqrt{x_{(q,q')}^2 + y_{(q,q')}^2} $ and $\nu = \text{atan} (\rho)$.  
%Thus, we could obtain the spherical convolutions for each point on the ERP.
% Besides, as the sampling locations are real-valued, we apply the bilinear interpolation
%, which is proved to be better than the other interpolations~\cite{coors2018spherenet}, 
% to the feature map.

\subsection{Reprojection Network}\label{sec:rep_repnet}
% TODO: 实验放在 supp 里面比较好
Though spherical convolution could reduce the distortion in ERP, the sampling locations of SphConv are biased when the network grows deeper~\cite{su2018kernel} such that the region proposed by SphRPN would be distorted on the sphere.
%the experiment in~\ref{} reveals that the sampling locations of the spherical convolutions are biased when the network grows deeper so that the regions proposed by SphRPN are distorted on the sphere.
Thus, we leverage the geometry of the spherical image and propose the reprojection network that takes the undistorted perspective projections as input.
% RepNet exploits a perspective projection based technique, \emph{i.e.}, Reprojection RoI pooling layer, to correct the proposed bounding boxes, and increases the accuracy of the detection.
%Thus, we propose the reprojection network, which leverages the geometry of the spherical image, and introduces a perspective projection base technique, \emph{i.e.}, Reprojection RoI pooling layer, to correct the proposed bounding boxes.
%Specifically, we extract a fixed-size feature for each spherical bounding box by the novel Reprojection ROI pooling layer, and then apply the regular Fast R-CNN to generate the relocated bounding boxes. 

\noindent\textbf{Reprojection RoI Pooling}:
Given the spherical bounding boxes proposed by SphRPN, reprojection RoI pooling layer generates the input of RepNet by transforming these spherical bounding boxes to fixed-length vectors.
%through expansion, reprojection and ROI pooling.
%The outputs of Reprojection RoI Pooling are fed into RepNet.
%Reprojection ROI pooling layer reprojects the bounding boxes to the tangent planes centered at the predicted object centers and resizes the obtained regions with the ROI pooling layer. 

Specifically, for each spherical bounding box $B^i$, since objects may be partially contained in $B^i$ due to the biased sampling locations, reprojection RoI pooling expands the FoVs of the spherical bounding box by a factor $r>1$, yielding a larger spherical bounding box $B_e^{i}=(B_{\theta}^i, B_{\phi}^i, rB_{fov_x}^i, rB_{fov_y}^i)$.
Then, the expanded spherical bounding box is reprojected to the tangent plane located at the predicted object center, \emph{i.e.}, $I_p[B_{\theta}^i, B_{\phi}^i]$. 
% TODO: 这里还是有点问题好吧
For each point $(\theta, \phi)$ within the spherical bounding box $B_e^{i}$, the corresponding coordinates in tangent plane $I_p[B_{\theta}^i, B_{\phi}^i]$ could be calculated by:
\begin{align}
\begin{small}
\begin{aligned}
	f_x(\theta, \phi) &= \frac{\cos\theta \, \sin(\phi - B_{\phi}^i)}{\sin B_{\theta}^i \, \sin\theta + \cos B_{\theta}^i \, \cos\theta \, \cos(\phi - B_{\phi}^i)}, \\
	f_y(\theta, \phi) &= \frac{\cos B_{\theta}^i \, \sin\theta - \sin B_{\theta}^i \, \cos\theta \, \cos(\phi - B_{\phi}^i)}{\sin B_{\theta}^i \, \sin\theta + \cos B_{\theta}^i \, \cos\theta \, \cos(\phi - B_{\phi}^i)}. 		
\end{aligned}
\end{small}
\end{align}
%The spherical bounding box corresponds to a rectangular region in the tangent plane.
%Since the spherical bounding box bounds a The bounding box corresponds to a rectangular region in the tangent plane.
Note that each spherical bounding box corresponds to a rectangular region in the tangent plane, while the shapes of the projected rectangular areas are various.
Thus, the reprojection RoI pooling exploits the RoI average pooling, which converts the projections into the patches with a fixed spatial extent of $H_p \times W_p$ (\emph{e.g.} $224 \times 224$).
%uses ROI average pooling to convert the various shapes of the rectangular regions on the tangent planes into the features with a fixed spatial extent of $H_p \times W_p$ (\emph{e.g.} $224 \times 224$).
%The RoI average pooling has the same effect as bilinear interpolation.

\subsection{Optimization}\label{sec:rep_optim}
\noindent\textbf{Loss Functions}:
We minimize a similar multi-task loss in both SphRPN and RepNet as Faster R-CNN~\cite{ren2015faster}.
Both networks have two sibling output layers~\cite{girshick2015fast}. 
Suppose that the concerned objects in the 360$^{\circ}$ images belong to a number of $K$
categories. 
The first layer outputs the probability distribution over $K+1$ categories (including background) by a softmax function, and the second layer outputs spherical bounding-box regression offsets parameterized by~\cite{girshick2014rich} for each object class.
The loss function for the spherical bounding-box regression $t$ and its objectness score $p$ is defined as:
\begin{equation}~\label{func:loss}
	L(p, t) = L_{cls}(p, p^*) + \lambda \, [p^* \geq 1] L_{reg}(t, t^*),
\end{equation}
where $p^*$ is the ground-truth label, and $t^*$ is the associated spherical bounding-box regression target. 
The classification loss $L_{cls}$ is the log loss for true class, while the regression loss $L_{reg}$ is the smooth $L_1$ loss defined in~\cite{girshick2015fast}.
The Iverson bracket indicator function [p$^* \geq$ 1] is applied to disable the regression loss for background (labeled by 0).

\noindent\textbf{SphRPN}:
We set $K=1$ in SphRPN, indicating whether the proposed regions belong to the foreground or background. 
Here, the references for the bounding-box regression are the default spherical anchors, which are assigned to foreground objects using a SphIoU threshold of 0.7, and to the background if the SphIoU is less than 0.3. We sample 128 positive and negative anchors per image with a ratio of 1:1.
The balance parameter $\lambda$ is set to 3 in all the experiments. 

\noindent\textbf{RepNet}:
Meanwhile, the number of categories $K$ is task-dependent in RepNet, and the references of bounding-box regression are the spherical bounding boxes generated by SphRPN. 
The spherical bounding box is now considered positive if the SphIoU between the prediction and the ground-truth box achieves at least 0.5, and negative if the SphIoU is less than 0.3. 
Besides, we sample 128 RoIs per image with a ratio of 1:3 of positive to negative, and set $\lambda=1$ in RepNet. 

%\noindent \textbf{Loss function}:
%We minimize the same multi-task loss in SphRPN as the Faster R-CNN~\cite{ren2015faster}. The loss function for a sampled anchor $A_i$ and its objectness score $p_i$ is defined as:
%\begin{equation}~\label{func:loss}
%	L(A_i, p_i) = L_{cls}(p_i, p_i^*) + \lambda \, p_i^* L_{reg}(A_i, A_i^*),
%\end{equation}
%
%Anchors are assigned to ground-truth object boxes using an SphIoU threshold of 0.7, and to background if the SphIoU is less than 0.3. The sampled positive and negative anchors have a ratio of up to 1:1.
%The balance parameter $\lambda$ is set to 3 in all the experiments. 
%
%
%\noindent \textbf{Loss function}:
%Identical with Fast R-CNN~\cite{girshick2015fast}, RepNet has two sibling output layers. 
%The first layer outputs the probability distribution over $K+1$ categories by a softmax function, and the second layer outputs bounding-box regression offsets for each of $K$ object classes.
%The loss function is similar with Eqn.~(\ref{func:loss}), except that the references are the spherical bounding boxes generated by SphRPN, rather than the default spherical anchors. 
%The spherical bounding box is now considered positive if the SphIoU between the prediction with the ground-truth box achieves at least 0.5, and negative if the SphIoU is less than 0.3. 
%Besides, we conduct negative sampling with pos-neg ratio as 1:2, and $\lambda=1$ in RepNet. 

\subsection{Implementation Details}\label{sec:rep_imple}
\noindent\textbf{Backbone}:
We apply VGG-16~\cite{simonyan2014very} as the backbone network for both stages. SphConv is adopted in the first stage, where conv5\_3 is served as the final feature map.
Since SphConv only changes the sampling locations in the convolutional filter, we could simply transfer the parameters between two VGG networks.
%initialize both models with the pre-trained model on ImageNet1k. 

\noindent\textbf{Anchors}: 
We use $k=9$ anchors in SphRPN.
Specifically, the anchors have three scales of $(30^{\circ})^2$, $(60^{\circ})^2$, and $(90^{\circ})^2$ with three aspect ratios of 1:1, 1:2, and 2:1.

\noindent\textbf{Training}:
%The ERP are resized to $512 \times 1024$, and the reprojection RoIs are resized to $224 \times 224$. 
%For the SphRPN, anchors are assigned to ground-truth object boxes using an IoU threshold of 0.7, and to background if the IoU is less than 0.3.
%While for the RepNet, an bounding box is considered positive if it has IoU with the ground-truth box of at least 0.5, and negative if the IoU is less than 0.3.
%The non-maximum suppression (NMS) with 0.7 threshold is adopted to reduce overlapping in SphRPN, and the top 300 proposal regions are extracted and fed to the RepNet.
%We randomly sample 256 anchors in an image to compute the loss function of a mini-batch, where the sampled positive and negative anchors have a ratio of up to 1:1
%In the first stage, we take the pos-neg ratio of 1:1, while in the second stage the pos-neg ratio is 1:2. 
We train Rep R-CNN in two steps. 
In the first step, we initialize SphRPN with the VGG-16 pre-trained on ImageNet dataset~\cite{russakovsky2015imagenet}, and fine-tune the network on specific 360$^{\circ}$ datasets with ERPs of size $512 \times 1024$ as input.
%based on the ERPs of size $512 \times 1024$.
In the second step, we adopt the regions proposed by SphRPN. 
The proposals are filtered by non-maximum suppression (NMS) with 0.7 threshold, reprojected to the tangent planes, and then resized to a fixed size of $224 \times 224$ 
as the input of RepNet.
We use SphRPN to initialize RepNet by duplicating the weights of SphConv directly to the normal convolutional filters. 
%The regions proposed by SphRPN are clipped by the non-maximum suppression (NMS) with 0.7 threshold, and the top 300 regions are fed into RepNet.
%RepNet using the proposals generated by the step-1 RPN. 
We do not share weights in backbone networks as we find that it would degrade the performance of the proposed detector.
Thus, we do not utilize alternating training \cite{ren2015faster} in Rep R-CNN.

Both SphRPN and RepNet are trained on 4 GPUs for 20 epochs. 
The batch size is set to 16 for SphRPN and 128 for RepNet.
The learning rate is initially set to 0.001 and then decreased by a factor of 10 after training 15 epochs. 
We use a stochastic gradient descent (SGD) optimizer with a weight decay of 0.0005 and a momentum of 0.9.
%Both loss functions in RPN and reprojection procedure are the same as~\cite{ren2015faster}, except that the width and height in the bounding-box regression are changed by FoVs following the spherical bounding box annotation.

\noindent\textbf{Inference}:
At test time, we propose the candidate regions based on the spherical anchors, and apply NMS with a threshold of 0.7 to reduce redundancy in SphRPN, which is the same as the training procedure.
After NMS, we use the top-$n$ ranked proposed regions for the second-stage detection.
We transform the selected proposals to fixed-size projections by reprojection RoI pooling, and then run RepNet on those projections, yielding rectified spherical bounding boxes as the output.
After that, we drop the proposals with less than 0.1 confidence score and apply NMS with a threshold of 0.45 to generate the final detections.
%We select top-$n$ (\emph{e.g.}, 50) proposed regions, and then transform these proposals with the Reprojection RoI pooling, yielding fixed size features as the input of the second stage.
%The number of proposal regions is set to $n$ (\emph{e.g.}, 50). 
%Afterwards, RepNet rectify the bounding boxes with RepNet, and clip the outputs with less than 0.1 confidence scores.
%Then, we run RepNet on the obtained projections, generating rectified bounding boxes, where the proposals with less than 0.1 confidence scores are dropped.
%After that, NMS with a threshold of 0.45 is applied to yield the final detections. 

\section{Experiments}\label{sec:experiment}
\subsection{Experimental Setup}
\noindent \textbf{Datasets}:
We evaluate the proposed Rep R-CNN on three datasets, including two novel synthetic datasets 
annotated by spherical bounding box and one real-world dataset without pre-labeled annotation.

\emph{VOC360}: 
VOC360 is a synthetic dataset generated from PASCAL VOC 2007 and 2012~\cite{everingham2010pascal} with 20 categories.  
We crop the objects with random-sized background from images in the VOC datasets, and then project the cropped images to arbitrary points on the sphere.
%The center of the object is aligned to the corresponding projection point, and the FoV angle of the longer side of each object is randomly selected from [60$^{\circ}$, 150$^{\circ}$].
Each image in VOC360 is attached by only one cropped image. 
VOC360 has 15000 training images, 1800 validation images, and 4955 test images.

\emph{COCO-Men}: 
%Similar with 
Referring to
the FlyingCars dataset~\cite{coors2018spherenet}, we construct the novel COCO-Men dataset.
COCO-Men combines the real-world background 360$^{\circ}$ images, and the segmented images of people cropped from COCO dataset~\cite{lin2014microsoft}. 
Each image includes three to six people, and every pair of people has an overlapping of SphIoU of less than 0.1.
%The FoVs of each people are chosen from [35$^{\circ}$, 70$^{\circ}$], which is different from VOC360.
%The larger FoV of each people is chosen from [35$^{\circ}$, 70$^{\circ}$], which is smaller than VOC360.
In total, the dataset comprises 4000 training images, 2000 validation images, and 1000 test images.

\emph{SUN360}:
To demonstrate the capability of Rep R-CNN for real scenes, we use SUN360 dataset~\cite{xiao2012recognizing} which contains a large number of real-world 360$^{\circ}$ images from the Internet.
In particular, we leverage the model trained on the VOC360 to examine the performance of Rep R-CNN on SUN360, regarding both accuracy and efficiency.

\noindent \textbf{Baseline Methods}:
We take the following state-of-the-art methods as baselines and compare the performance of Rep R-CNN with the baseline methods.
% We compare Rep R-CNN to the following state-of-the-art methods. 
Note that all baseline methods are one-stage object detectors.

\emph{ERP-SSD}~\cite{liu2016ssd}: 
We apply SSD directly to spherical images represented by ERP.

\emph{Multi-projection}:
%We perform multiple perspective projections on the sphere, where the projection points have equal intervals.
We select overlapping projection areas with equal intervals to cover the sphere, and then perform perspective projection on these areas.
The projections are then fed into a Fast R-CNN.
%A normal CNN is applied on the locations and outputs the classification score bounding box regression of each projection.

\emph{Sphere-SSD}~\cite{coors2018spherenet}:
The Sphere-SSD is constructed by replacing normal convolutions in ERP-SSD with SphConv.

\emph{SPHCNN}~\cite{su2017learning}:
%We modify the architecture of the authors' implementation.
%, since the original object detector is not trainable. 
%We project the conv5\_3 feature map to the tangent plane following~\cite{su2017learning}, and then add a $1 \times 1$ conv to generate both bounding box regressions and classification scores.
We add a $1 \times 1$ conv on top of the conv5\_3 feature map~\cite{su2017learning} based on the author's implementation to generate spherical bounding-box regressions and classification scores for 
%object detection
an actual 360$^{\circ} $object detection.
%Since there are numerous parameters in the network, we only train the network on top of conv5\_3 that do not received the knowledge transfer.
%We train the modified network through the multi-task loss in RepNet.

\emph{S$^2$CNN}~\cite{cohen2018spherical}: 
%We train S$^2$CNN using the authors' implementation. 
Since S$^2$CNN is originally designed for classification, we adapt S$^2$CNN to object detection.
%We select several regions on the sphere to cover the image, and apply perspective projection on the selected regions.
We use multiple perspective projections to represent the 360$^{\circ}$ images.
Then, we feed the projections into the authors' implementation of S$^2$CNN, and concatenate the outputs of the network.
We propose regions on top of the combined feature map.
To avoid out-of-memory error, we scale down the input resolution to $64 \times 64$ suggested by authors.

\emph{Spherical CNN}~\cite{esteves2018learning}:
We modify the authors' implementation in the same way as S$^2$CNN.
The input is again scaled down to $64 \times 64$ due to the memory limit.

For a fair comparison, the baseline methods are tuned by either implementation with recommended parameter settings or grid search for the best performance. 
In addition, the backbone networks are all the same VGG-16 except for S$^2$CNN and Spherical CNN. 
Please refer to supplementary material for additional details of datasets and baseline methods.

\noindent \textbf{Performance Metric}: 
For the performance measure of the 360$^{\circ}$ object detection, we adopt the standard Average Precision (AP) for each individual object class, and report the mean Average Precision (mAP) for all classes \cite{everingham2010pascal}. 
A detection is considered to be correct when the IoU between the prediction and ground-truth exceeds 50\%.
Here, we use the actual IoU calculated by the pixel-level integral on the sphere to ensure that the measurement is unbiased and fair to all comparison algorithms.

\begin{table}[tp]
\centering
\resizebox{0.95\linewidth}{!}{
\begin{tabular}{c|c|c|c|c}
& $n$ 		& VOC360 	& COCO-Men & Speed      \\                                                        \midrule[1pt]
ERP-SSD~\cite{liu2016ssd}		& -		& 32.69		& 44.20		& 76ms\\
Multi-projection	& 200	& 49.16		& 62.35		& 273ms	\\
Sphere-SSD~\cite{coors2018spherenet}			& -		& 48.25		& 54.79		& 86ms	\\
SPHCNN~\cite{su2017learning}				& - 		& 49.41		& 48.17		& 224ms	\\
S$^2$CNN~\cite{cohen2018spherical}			& 50		& 37.45		& 45.36		& 139ms\\
Spherical CNN~\cite{esteves2018learning}		& 50		& 35.12		& 41.53		& 145ms	\\
\hline
Rep R-CNN	& 10		& 69.70		& -		& 112ms	\\
Rep R-CNN	& 20		& \textbf{71.88}		& 74.72		& 127ms	\\
Rep R-CNN	& 50		& 71.65		& \textbf{81.48}		& 178ms\\
Rep R-CNN	& 100	& -			& 81.34		& - \\
\end{tabular}
}
\centering
\caption{Performance comparison in terms of mAP on both VOC360 and COCO-Men datasets and speed (ms per image). In the baseline methods, proposal represents the number of perspective projections fed to the networks; while in Reprojection R-CNN, it represents the number of proposals fed to RepNet. The boldface denotes the best performance on each dataset.}\label{tab:baseline}
\end{table}

%\begin{figure}[tb]
%    \centering
%    \includegraphics[width=0.75\linewidth]{fig/cur0.pdf}
%    \caption{Polar angle/mAP curves of Rep R-CNN and three baseline models.}~\label{fig:cur0}
%\end{figure}

%\begin{figure}[tb]
%    \centering
%    \includegraphics[width=0.75\linewidth]{fig/cur1.pdf}
%    \caption{Polar angle/mAP curves of SphRPN and RPN.}~\label{fig:cur1}
%\end{figure}

\begin{figure*}[tp]
\begin{minipage}[t]{0.32\linewidth}
\centering
\includegraphics[width=\textwidth]{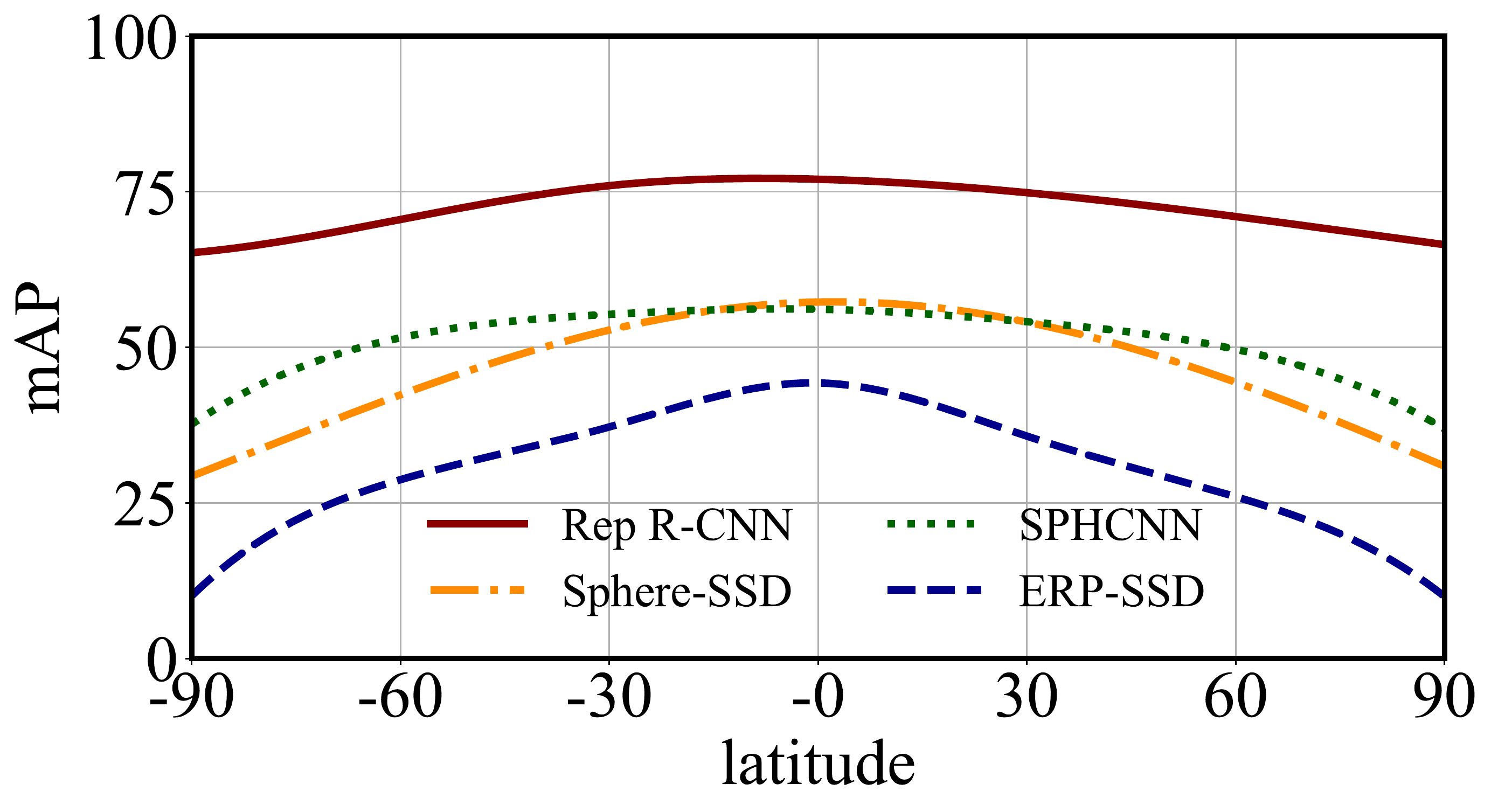}
\caption{Polar angle/mAP curves of Rep R-CNN and three baselines on VOC360.}
\label{fig:cur0}
\end{minipage}
\hfill
\begin{minipage}[t]{0.32\linewidth}
\centering
\includegraphics[width=\textwidth]{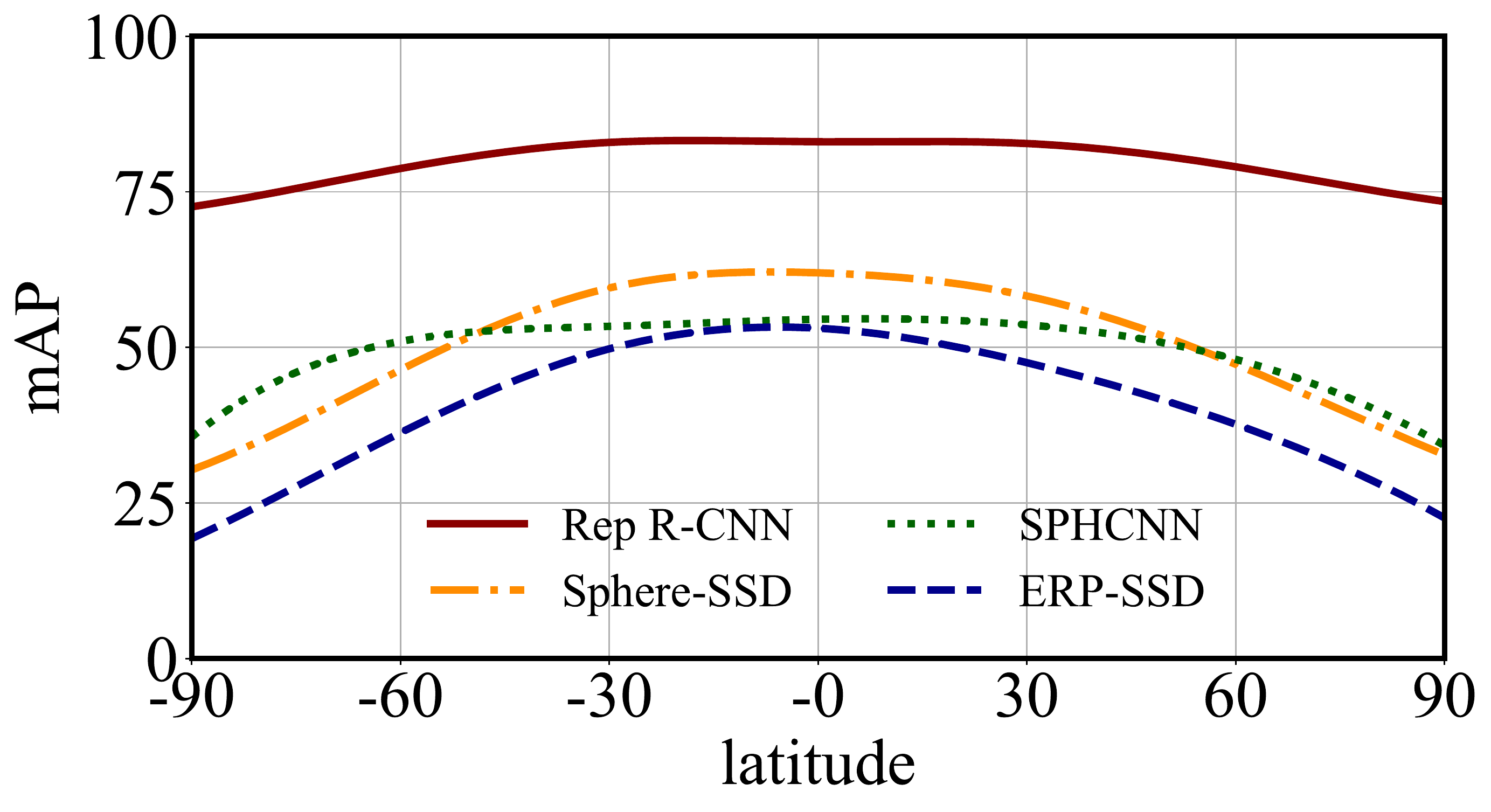}
\caption{Polar angle/mAP curves of Rep R-CNN and three baselines on COCO-Men. }
\label{fig:cur2}
\end{minipage}
\hfill
\begin{minipage}[t]{0.32\linewidth}
\centering
\includegraphics[width=\textwidth]{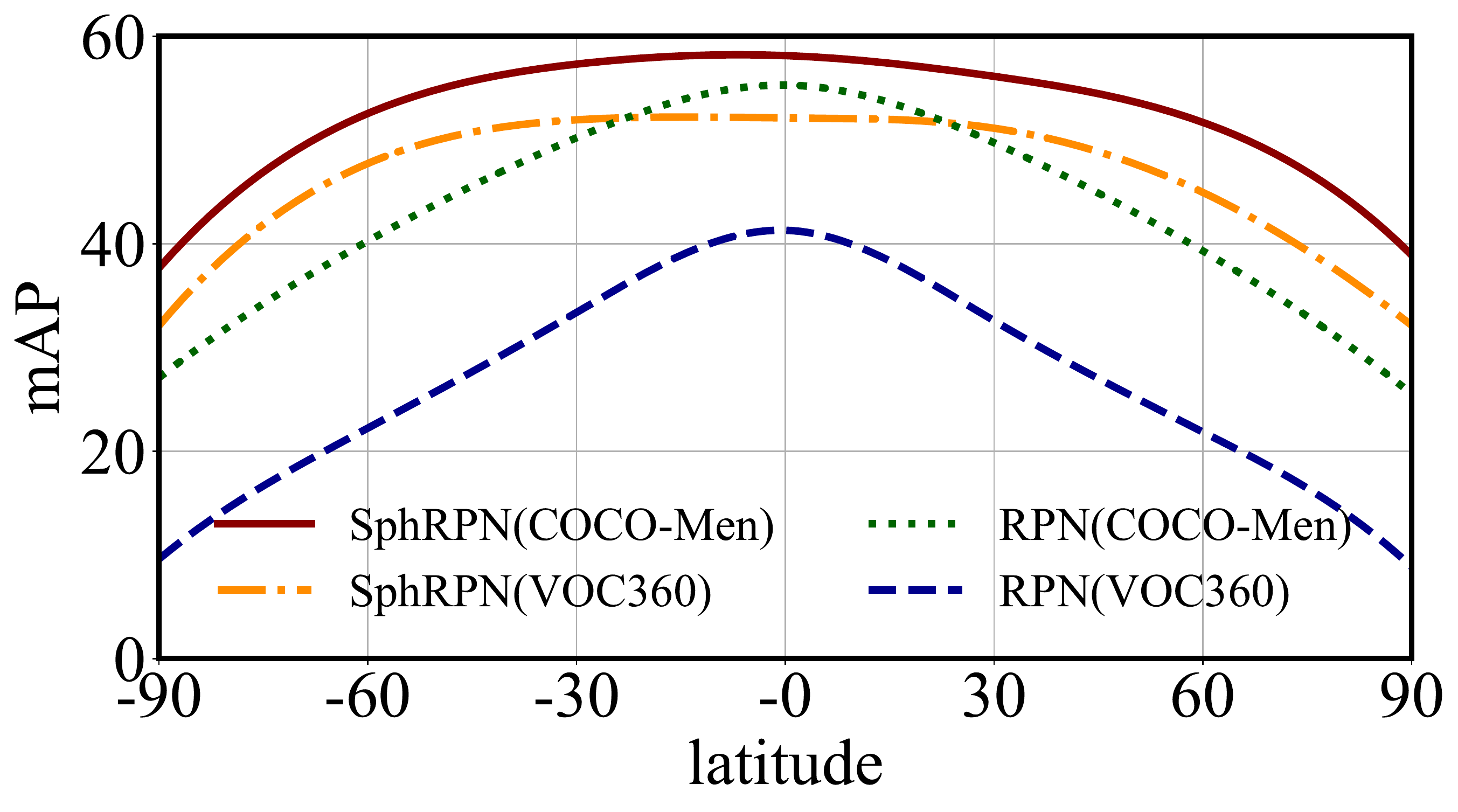}
\caption{Polar angle/mAP curves of SphRPN and RPN on both datasets.}
\label{fig:cur1}
\end{minipage}
\end{figure*}

\subsection{Performance of Rep R-CNN}
We compare the proposed Rep R-CNN with the baseline methods in both VOC360 and COCO-Men datasets.
The results are shown in Table~\ref{tab:baseline}.
Among the baseline methods, it is obvious that Multi-projection achieves relatively good performance but is time-consuming due to the large number of region proposals. 
Owing to the use of SphConv, Sphere-SSD exhibits competitive performance in both datasets with almost 3x speed faster than the Multi-projection method.
Besides, the other novel methods for 360$^{\circ}$ images, \emph{i.e.}, SPHCNN, S$^2$CNN and Spherical CNN show less competitive performance than the above two methods in either speed or accuracy due to the memory constraints.
%since dense region proposals would result in out of memory errors in S$^2$CNN and Spherical CNN and numerous parameters in SPHCNN, the performance of the three methods are less competitive in either speed or accuracy.

Regarding Rep R-CNN, 
%it is obvious 
it can be observed 
that the proposed detector adopts the rapid and relatively precise SphConv in the region proposal network, and then regresses the proposed regions with the accurate perspective-projection based method.
Therefore, Rep R-CNN can be regarded as a combination of Sphere-SSD and Multi-projection, which is both fast and accurate.
The results in VOC360 and COCO-Men convincingly demonstrate the effectiveness of the proposed method.
%The proposed Rep R-CNN largely outperforms other methods on both VOC360 and COCO-Men, which convincingly demonstrates the effectiveness of the proposed \textbf{two-stage} method. 
Specifically, Rep R-CNN achieves 71.88 mAP on the VOC360 dataset, exceeding the strongest baseline, \emph{i.e.}, SPHCNN, by over 45\%; while the performance gain in the COCO-Men dataset is also at least 30\% compared to the previous state-of-the-art Multi-projection method.
In addition, Rep R-CNN achieves the best performance with only 127ms per image on VOC360 dataset, which is faster than almost all the baseline methods. 
%Besides, Rep R-CNN could also run at 178ms per image with 50 region proposals, demonstrating its efficiency for real-world 360$^{\circ}$ tasks.

\begin{figure*}[t]
    \centering
    \includegraphics[width=\linewidth]{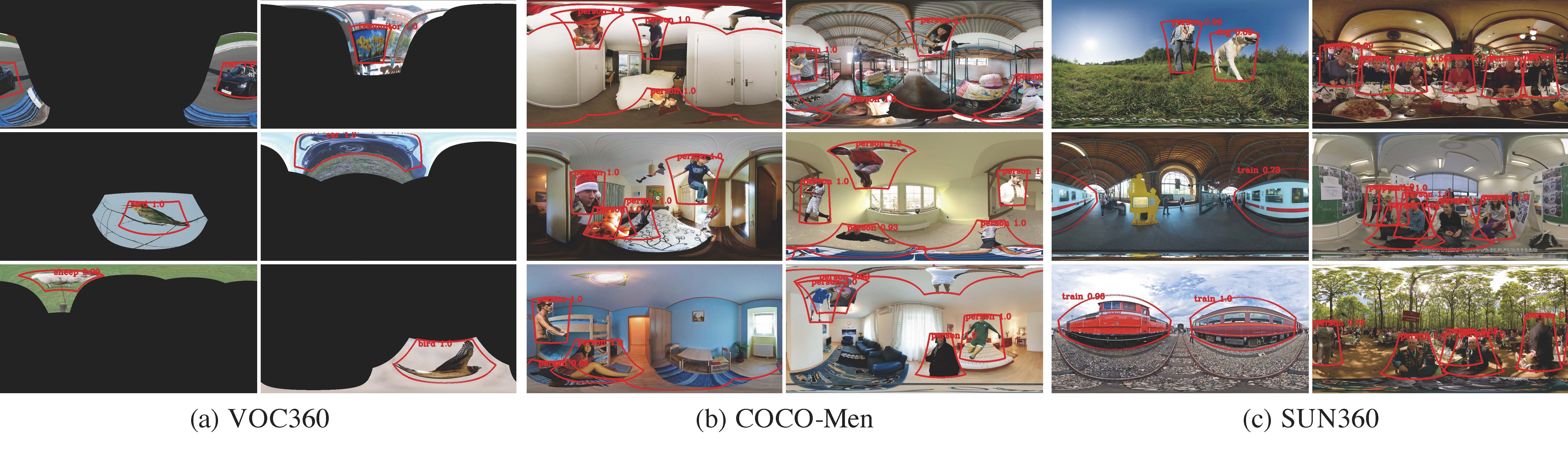}
    \caption{More results of Rep R-CNN on the three datasets.}
    \label{fig:casestudy}
\end{figure*}

% TODO:稍微挑一个结果图
Moreover, to show the robustness of Rep R-CNN, we examine the mAP of the detection algorithms by varying the polar angle, and plot the polar angle/mAP curves of methods in Figure~\ref{fig:cur0} and~\ref{fig:cur2}.
%Specifically, since the perspective projection based methods are not affected by polar angle, we only consider the methods that take ERP as input.
Since the perspective-projection based methods are not affected by polar angle, we only consider the methods that take ERP as input.
We find that Rep R-CNN forms an upper envelope over all existing methods.
Furthermore, though the distortion in ERP varies with the polar angle, Rep R-CNN is only slightly affected, and exhibits competitive performance even if the objects are extremely distorted, \emph{i.e.}, near the poles.
Rep R-CNN outputs are visualized in Figure~\ref{fig:casestudy}.
The result reveals that the proposed Rep R-CNN is robust to the various distortion and discontinuity in 360$^{\circ}$ images.

\begin{table}[tp]
\centering
\begin{small}
\begin{tabular}{c|c|c}
	& VOC360 	& COCO-Men       \\                                                        
\midrule[1pt]
RPN				& 30.91		& 42.79	\\
SphRPN			& 44.71		& 50.26	\\
\hline
RPN + RepNet	& 63.82		& 72.81	\\
Rep R-CNN		& \textbf{71.88}		& \textbf{81.48}	\\
\end{tabular}
\end{small}
\centering
\caption{Performance comparison of RPN, SphRPN, RPN + RepNet and Rep R-CNN on both synthetic datasets. We employ the same settings in normal RPN and SphRPN, including network architecture, hyper-parameters, and the second-stage RepNet.}\label{tab:SphConv}
\end{table}

\subsection{Ablation Experiments}
We conduct several ablation studies to verify the design of Rep R-CNN.

\noindent\textbf{Region Proposal Network}:
The previous object detectors for 360$^{\circ}$ images are all one-stage, and neglect the process of region proposal.
To demonstrate the necessity of RPN, we compare Rep R-CNN with Multi-projection,
because both methods utilize perspective projections as input and the only difference
% where both methods utilize perspective projections as input.
%The only difference 
is whether the projections are generated by RPN or by uniform sampling.
For a fair comparison, we 
%apply the same VGG-16 network in both Multi-projection and RepNet, and 
generate about 200 projection areas for Multi-projection such that the proposals could cover all the objects on the sphere compactly.
As shown in Table~\ref{tab:baseline}, Multi-projection is inferior to Rep R-CNN, especially in the VOC360 dataset where the objects are small and difficult to locate.
Moreover, in order to get exact proposals, Multi-projection method always samples numerous candidate projections, leading to large time consumption.
Therefore, we could infer that a preceding region proposal process is crucial for the fast and accurate object detector in 360$^{\circ}$ images.

\noindent\textbf{Reprojection Technique}:
%We introduce the novel reprojection technique in Rep R-CNN.
We then consider the efficiency of reprojection technique. 
In this experiment, we take Sphere-SSD as the comparison since Sphere-SSD is a variant of SphRPN, which can be regarded as a Rep R-CNN without RepNet.
%The results are shown in Table~\ref{}.
% test the performance of a single SphRPN, and report the result in Table~\ref{}.
It can be observed from Table~\ref{tab:baseline} that the lack of RepNet leads to a substantial decline of more than 30\% in mAP, indicating that the strength of Rep R-CNN is mainly attributed to the reprojection-based bounding-box relocation.
Moreover, we consider the mAP at different latitudes. 
As shown in Figure~\ref{fig:cur0} and~\ref{fig:cur2}, although Sphere-SSD fits the sampling locations of the convolutional kernel on the sphere, it still suffers from distortion in the polar regions.
% and captures approximate location of the objects.
In contrast, Rep R-CNN is capable of detecting and localizing the objects at any location, indicating that the reprojection technique could adjust the bias introduced by SphRPN.

%including the boundary and the polar regions.
 
\noindent\textbf{Spherical Convolution}:
% TODO：可以加一些图，表明这个 feature map 代表的特征位置就不正确，比如说第一阶段就框的位置差很远，这样第二阶段也不能回归出结果，因为根本看不出结果
%owing to the corrected sampling locations of SphConv, justifying the use of SphConv in our region proposal network. 
In accordance with the result in~\cite{coors2018spherenet}, we find that 
%owing to SphConv, 
Sphere-SSD performs better than vanilla SSD in both datasets as shown in Table~\ref{tab:baseline}.
In this section, we investigate the effect of SphConv in the two-stage detector.
We replace SphConv with the normal convolution in RPN, and report the result of the modified Rep R-CNN as well as RPNs in Table~\ref{tab:SphConv}. 
%Similar with the result between Sphere-SSD and normal SSD, 
Similar to the previous outcomes, SphRPN outperforms the normal RPN in both datasets.
Though the second-stage relocation fills the gap between SphRPN and normal RPN,
% between Rep R-CNN and the modified Rep R-CNN
this alternate still results in a mAP loss of nearly 14 points in VOC360 dataset and over 7 points in COCO-Men dataset, implying that the distortion-aware SphConv is crucial for the two-stage detector in 360$^{\circ}$ images.
Besides, while the accuracy degrades, the proposed two-stage detector still largely surpasses the state-of-the-art methods with the normal convolution, which once again highlights that reprojection-based second stage is the core of the proposed Rep R-CNN.

To give a detailed explanation of SphConv, we plot the polar angle/mAP curves of SphRPN and the normal RPN in Figure~\ref{fig:cur1}.  
Though SphConv proposes coarse spherical bounding-box regressions, it is less impacted by the distortion near the poles.
In comparison, normal RPN fails to detect the objects apart from the equator or near the prime meridian, and thus leading to a drop in the accuracy.

%Moreover, since RepNet demands for approximate locations of objects to crop the proposal regions, .
%On the other hand, SphRPN could also fail at the xxx, which once again indicates the rationality of the two-stage method. 

\noindent\textbf{Number of Region Proposals}:
%An evaluation of the proposed reprojection technique is shown in Table~\ref{}.
To balance the speed and accuracy, we compare a different number of region proposals during the inference time.
In Table~\ref{tab:baseline}, we observe that Rep R-CNN could achieve the best mAP with only 20 proposals in VOC360, and 50 proposals in COCO-Men dataset, which is much less than the Multi-projection with 200 proposals.
The result indicates that SphRPN can accurately generate the region of interests in the ERP while the second-stage RepNet can extract specific objects from the proposed regions.
%only few region proposals could accurately locate to the region of interests.
Additionally, with 50 proposals, Rep R-CNN could run at 178ms per image on an NVIDIA Tesla V100 GPU, which means the proposed algorithm could be applied in VR and real-world omnidirectional video streaming.
%Additionally, the proposals more than 30 could lead to degradation in the performance, .

\subsection{Rep R-CNN in Real-world Dataset}
To verify that the proposed Rep R-CNN is also effective in real-world scenarios, we exploit the network trained on VOC360, and directly apply it to the real-world SUN360 dataset without fine-tuning.
%Figure~\ref{fig:casestudy} shows several object detection examples on the three datasets. 
As shown in Figure~\ref{fig:casestudy}.c), consistent with the previous experiments, objects of various categories at different latitudes can be successfully detected by the proposed Rep R-CNN despite the distortion and discontinuity.
This result demonstrates that Rep R-CNN could perform well over the real-world scenarios.
%Rep R-CNN could locate objects of various categories at different latitudes on the sphere.
%In particular, the proposed detector successfully detects objects despite the distortion and discontinuity, demonstrating that Rep R-CNN could perform well over the real-world scenarios.
%, where we observe a good performance on the VOC360 dataset and the COCO-Men dataset. 
%These models transfer the knowledge and perform well over the SUN360 dataset on both the multi-class detection task (as shown in the first column of Figure~\ref{fig:casestudy}.c) and the single-class detection task (as shown in the second column of Figure~\ref{fig:casestudy}.c).
%Consistent with the previous experiments, Rep R-CNN could detect objects at different latitudes on the sphere, including heavily-distorted polar regions and boundary regions where regular methods fail.
Additional qualitative experiments of Rep R-CNN are provided in the supplementary material.

\section{Conclusion}
%Along with the broad spread of omnidirectional applications, object detection in 360$^{\circ}$ images has attracted much attention from engineers and researchers.
In this work, we present a standard framework to address the object detection for 360$^{\circ}$ images, including novel spherical criteria and a two-stage object detector named Rep R-CNN.
%To address the distortion of objects and the lack of appropriate annotations of the 360$^{\circ}$ images, we present a standard object detection framework for 360$^{\circ}$ images.
%Within the presented framework, we propose a novel two-stage object detector named Rep R-CNN.
Rep R-CNN combines the strength of both ERP and perspective projection, resulting in fast and accurate object detection in 360$^{\circ}$ images.
%The first stage generates coarse region proposals by a distortion-aware SphRPN over the entire ERP images.
%The second stage concentrates on the undistorted perspective projection and processes the proposed regions by RepNet.
%A Reprojection pooling layer is applied to coordinate the two stages and convert the projections into the features with a fixed spatial extent.
We introduce two synthetic 360$^{\circ}$ image datasets to examine the performance of Rep R-CNN.
%We prepare two synthetic 360$^{\circ}$ image datasets and one real-world 360$^{\circ}$ image dataset to examine the performance of Rep R-CNN.
Experimental results show that Rep R-CNN outperforms several state-of-the-art 360$^{\circ}$ object detectors by at least 30\% at high speed.
Ablation experiments also verify the efficiency and accuracy of Rep R-CNN.
%, indicating that Rep R-CNN is applicable to the real-world omnidirectional applications.
In addition, the model can be transferred to the real-world dataset and remain good performance, indicating that Rep R-CNN is applicable to the real-world scenarios.
%on the multi-class multi-object detection task.

\balance
{\small
\bibliographystyle{ieee}
\bibliography{RepRCNN}
}

\end{document}